\documentclass[10pt,twocolumn,letterpaper]{article}

\usepackage[pagenumbers]{cvpr} 

\definecolor{cvprblue}{rgb}{0.21,0.49,0.74}
\usepackage[pagebackref,breaklinks,colorlinks,allcolors=cvprblue]{hyperref}
\usepackage{verbatim}
\usepackage{soul}
\usepackage{array}
\usepackage{xcolor,colortbl}

\usepackage{tikz}
\usepackage{adjustbox}

\newcommand{\closeruline}[1]{%
  \begin{tikzpicture}[baseline=(textnode.base)]
    \node[inner sep=0.5pt,outer sep=0.5pt] (textnode) {#1};
    \draw[yshift=0ex] (textnode.south west) -- (textnode.south east);
  \end{tikzpicture}%
}

\usepackage{float} 

\def\paperID{7080} 
\def\confName{CVPR}
\def\confYear{2025}

\title{LinVT: Empower Your Image-level Large Language Model\\to Understand Videos}

\author{Lishuai Gao, Yujie Zhong$^{\dagger}$, Yingsen Zeng, Haoxian Tan, Dengjie Li, Zheng Zhao \\
Meituan Inc.\\
{\tt\small gaolishuai0425@163.com,} 
{\tt\small jaszhong@hotmail.com} 
}

\usepackage{multirow}
\begin{document}

\maketitle

\renewcommand{\thefootnote}{\fnsymbol{footnote}}
\footnotetext{$^{\dagger}$Corresponding author.}

\begin{abstract}
Large Language Models (LLMs) have been widely used in various tasks, motivating us to develop an LLM-based assistant for videos. Instead of training from scratch, we propose a module to transform arbitrary well-trained image-based LLMs into video-LLMs (after being trained on video data). To better adapt image-LLMs for processing videos, we introduce two design principles: linear transformation to preserve the original visual-language alignment and representative information condensation from redundant video content. Guided by these principles, we propose a plug-and-play Linear Video Tokenizer (LinVT), which enables existing image-LLMs to understand videos. 
We benchmark LinVT with six recent visual LLMs: 
Aquila, Blip-3, InternVL2, Mipha, Molmo and Qwen2-VL, showcasing the high compatibility of LinVT.
LinVT-based LLMs achieve state-of-the-art performance across various video benchmarks, illustrating the effectiveness of LinVT in multi-modal video understanding.
Code is available at \href{https://github.com/gls0425/LinVT}{https://github.com/gls0425/LinVT}.
\end{abstract}    
\section{Introduction}
\label{sec:intro}

\begin{figure}[htbp]
    \centering
    \begin{subfigure}[b]{\columnwidth}
        \centering
        \includegraphics[width=1.0\columnwidth]{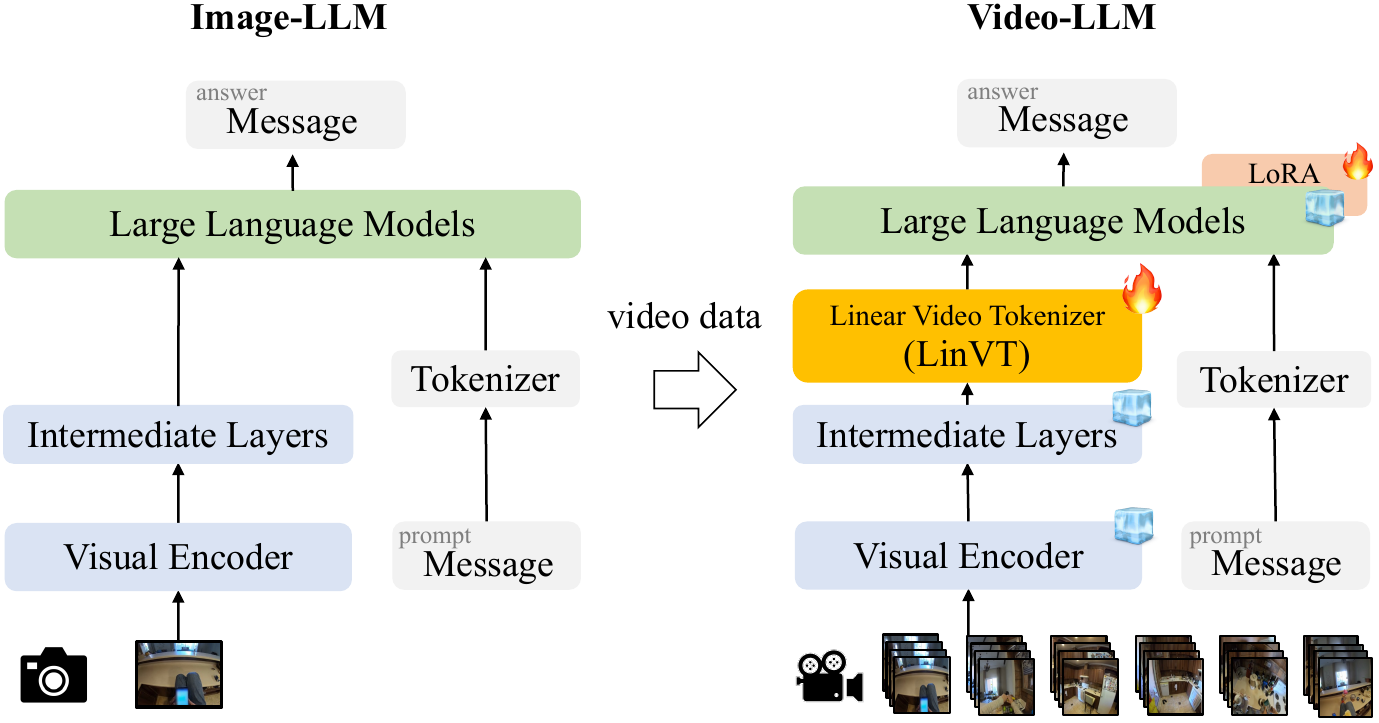}
        \caption{LinVT can convert an image-LLM to a video-LLM, with high compatibility. 
        The intermediate layers vary across different image-LLMs. For instance, it denotes a MLP followed by a pooling layer in Molmo~\cite{deitke2024molmo}, or a token sampler in BLIP-3~\cite{xue2024xgen}. }
        \vspace{1mm}
        \label{fig:ffc_pipeline}
    \end{subfigure}
    \vspace{0.8em}
    \begin{subfigure}[b]{\columnwidth}
        \centering
        \includegraphics[width=0.8\columnwidth]{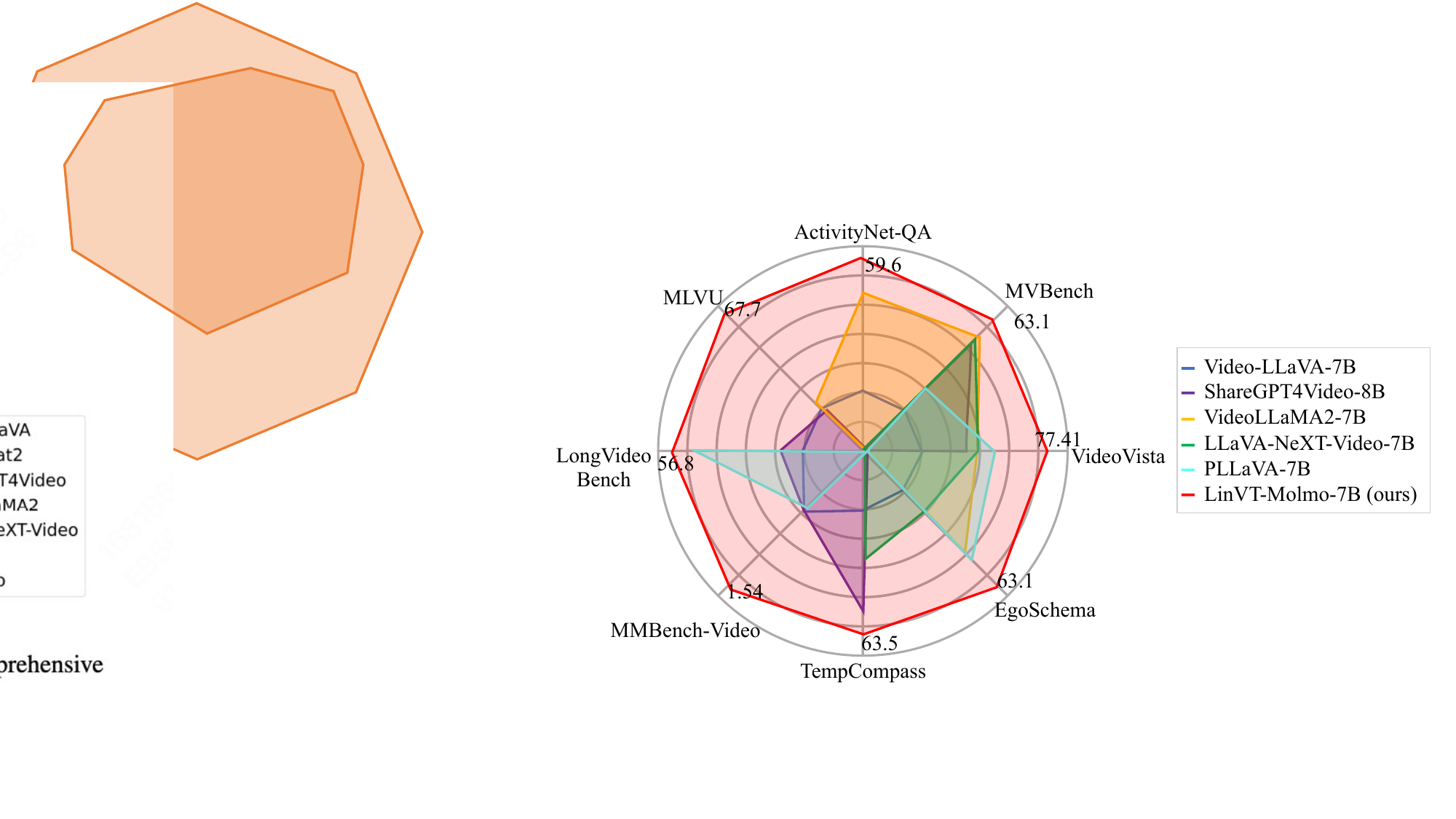}
        \caption{Comparison between LinVT-Molmo-7B (ours) and other video-LLMs. }
        \label{fig:perform}
    \end{subfigure}
    \caption{By being trained on video data, LinVT can endow an image-based LLM with the capability to handle video understanding tasks and achieve outstanding performance.}
    \label{fig:intro_fig}
    \vspace{-2mm}
\end{figure}

The rapid proliferation of video data has spurred extensive research in the fields of computer vision and natural language processing, aiming at the effective comprehension and processing of lengthy video content. Long videos inherently encompass a wealth of temporal and spatial information, making the comprehension process remarkably demanding. Consequently, the scientific community has been devoted to developing intelligent data assistants capable of time-sensitive analysis of prolonged videos, including tasks such as 
dense video captioning, 
temporal grounding, key moment summarization, \etc. Large language models (LLMs) have consistently demonstrated exceptional comprehension capabilities, thus motivating our endeavor to develop an LLM-based assistant that facilitates a comprehensive understanding of long videos.

Due to the large-scale training of image-text pairs, large-scale image-LLMs \cite{chiang2023vicuna,liu2024visual,li2023blip,chen2024far,zhu2023minigpt} have become dominant in computer vision.
The development of image-LLMs is progressing rapidly, showcasing numerous improvements. Meanwhile, many efforts are struggling to train video-LLMs from scratch using vast amounts of image and video training data. 
For instance, Qwen-VL~\cite{bai2023qwen} requires over one billion 
images
for training, and Blip-3~\cite{xue2024xgen} even requires over one trillion.
When training a video-LLM, a natural strategy is to build on existing image-LLMs due to their powerful image-level comprehension abilities, which also form the basis of understanding videos. This approach allows us to avoid training a video-LLM from scratch, thereby saving extensive computational resources and time. Consequently, developing a method to leverage the capabilities of sophisticated image-LLMs for training video-LLMs would be highly valuable in practice.
To achieve this, we develop a video tokenizer that can be injected into image-LLMs, and its aim is to transform the visual tokens (of the video frames) into meaningful video tokens that can be understood by the LLMs, as illustrated by Fig.~\ref{fig:ffc_pipeline}.

In designing the video tokenizer, an important principle is to ensure that the video-LLM retains the image-level knowledge both during and after training on video data.
Notably, preserving image-level comprehension ability within the video-LLM is crucial, as it allows for a single model to be used, leading to a simpler architecture and facilitating easier deployment for practical applications.
In this work, we posit that a key to satisfy this requirement is \emph{to maintain the visual-language alignment of the image-LLM in the video-LLM.}
Therefore, our first design principle of the video tokenizer is that its output should be the \textbf{linear combination} of its input.

Apart from the linearity design principle, another crucial goal of the video tokenizer is the information refinement. 
An obvious distinction between images and videos is that videos may contain a great many of frames (especially for long videos), thereby producing significantly more visual tokens which are fed to the LLM. In this case, excessive visual tokens can lead to a computationally overwhelming burden, as well as being at rish of exceeding the context window of LLMs. Nevertheless, a sophisticated video tokenizer should have the capability to extract reasonable number of information-rich visual tokens from the redundant video content. We coin this design principle as \textbf{representative information condensation}.

Previous video-LLMs employ token compression techniques in tackling the information-redundancy issue, such as Video-LLaMA~\cite{zhang2023video}, VideoLLaMA2~\cite{cheng2024videollama}, PLLaVA~\cite{xu2024pllava}. 
However, they fail to take into account two important perspectives: tackling events of different temporal lengths (\ie duration) and dynamically extracting question-related information (provided by the users) from the video. 
Therefore, in this work, we wish to strengthen the ability of the video tokenizer in extracting representative information in these two essential aspects.

Following the two abovementioned design principles, we propose a module termed Linear Video Tokenizer (LinVT) as depicted in ~\cref{fig:ffc_pipeline}. LinVT is positioned right before the LLM. It takes a sequence of image-level tokens as input and generates video-level visual tokens. 
The output video tokens of LinVT are strictly the weighted average of (part of) the input visual tokens of LinVT, ensuring that the knowledge from the image-LLM is effectively preserved. 
Furthermore, LinVT integrates multi-scale processing of the visual tokens, generating token sequences of varying lengths to accommodate videos (and events) of different durations. Lastly, by involving text interaction, LinVT is able to accurately extract question-relevant information from the video.

Extensive experiments demonstrate the effectiveness of LinVT.
LinVT is a plug-and-play module and we combine it with six recent multi-modal LLMs, namely, 
Aquila~\cite{gu2024infinitymmscalingmultimodalperformance}, Blip-3~\cite{xue2024xgen}, InternVL2~\cite{chen2024far}, Mipha~\cite{zhu2024comprehensive}, Molmo~\cite{deitke2024molmo} and Qwen2-VL~\cite{wang2024qwen2}, 
for video understanding tasks in the experiments. 
In particular, Blip-3, Mipha and Molmo are three image-LLMs that are not trained on video data, whereas Aquila, InternVL2 and Qwen2-VL are trained on image data as well as video data. Not surprisingly, LinVT brings significant video understanding ability to all of them. Fig.~\ref{fig:perform} shows the performance of LinVT-Molmo-7B.
More impressively, LinVT-based video-LLMs achieve superior performance with high training efficiency, namely utilizing \textbf{only video data} (without any image data) for training.

To summarize, our contributions are listed as follows:

\begin{itemize}[leftmargin=0.5cm]

\item 
We propose Linear Video Tokenizer (LinVT), a module that enables image-LLMs to understand videos. In particular, the output of this tokenizer is strictly the weighted average of (part of) the input frame-level visual tokens, ensuring that the knowledge from the image-LLM is effectively preserved.

\item 
Based on the principles of representative information condensation, LinVT integrates multi-scale processing to tackle events of various durations. Furthermore, it involves text conditions to extract question-relevant visual cues from videos.

\item 
Extensive experiments demonstrate the effectiveness of LinVT. We inject LinVT into six different types of image-LLMs, and all of them achieve outstanding performance in video understanding tasks.

\end{itemize}
\section{Related Work}
\label{sec:related}

\textbf{Video Large Language Models.}~
Inspired by the powerful understanding capabilities and world knowledge of Large-scale Language Models (LLMs), many efforts~\cite{liu2024visual,wang2024qwen2,chen2024far,li2023blip,alayrac2022flamingo} have been devoted to integrating textual and visual information multi-modal tasks. 
Recent works have achieved success in combining LLMs with video encoders, which allows for the powerful understanding and generation capabilities of LLMs to address video-related tasks. 
The primary distinction among these multi-modal LLMs lies in the way they encode videos into visual tokens. In notable works like VideoChat~\cite{li2023videochat}, video backbones are employed to encode video embeddings, followed by the use of a query Transformer, \ie Q-Former~\cite{instructblip}, to compress video tokens. Conversely, Video-LLaMA~\cite{zhang2023video} first utilizes a visual Transformer (ViT) and an image Q-Former to encode individual frames and then incorporates a video Q-Former for temporal modeling. 
However, the methods suffer from a limitation as they can only handle a limited number of video frames due to a large number of video tokens produced by the encoder. This constraint ultimately leads to a loss of visual semantics when processing long videos.
In our work, we propose a module specifically designed for representative token condensation in videos.

\noindent \textbf{Vision-language instruction tuning.}~
Inspired by the recent success of fine-tuning in LLMs, visual language instruction tuning is often incorporated after the visual-text alignment stage. The goal of instruction tuning is to enhance the instruction-following capabilities. 
This tuning requires generating high-quality data through human-guided instructions, which can be categorized into two technical branches. The first branch integrates existing multi-modal task datasets and converts them into instruction formats, as exemplified by frameworks such as MultiInstruct~\cite{xu2022multiinstruct}, InstructBLIP~\cite{instructblip}, and M$^3$IT~\cite{li2023m3it}. 
The second approach leverages GPT-4~\cite{openai2023gpt} to generate more diverse conversational data. Methods such as miniGPT4~\cite{zhu2023minigpt}, LLaVA~\cite{liu2024visual}, MIMIC-IT~\cite{li2023mimic}, VideoChat~\cite{li2023videochat}, and Valley~\cite{luo2023valley} focus on obtaining detailed visual descriptions and constructing image or video-based dialogue data from LLMs. However, these methods tend to overlook the temporal dimension of user requests in relation to video understanding. To address this, several datasets \cite{maaz2023video,caba2015activitynet,xiao2024can,liu2024bench,patraucean2024perception} focusing on video temporal understanding have been developed.

\noindent \textbf{Visual token pruning.}~
Visual token pruning is widely adopted to reduce redundancy and overlapping information in vision encoders and LLMs while retaining task-relevant information. 
\cite{ren2023testa} introduced a temporal aggregation module to combine duplicate video frames and a spatial aggregation module to merge similar image blocks within each frame.
\cite{shen2024tempme} addressed temporal redundancy by progressively merging tokens in adjacent segments, thereby reducing the number of tokens while preserving crucial video-level features. 
\cite{shang2024llava} introduced adaptive token reduction techniques involving adaptive token selection and token supplementation, which can be employed without fine-tuning. 
In LLMs, pruning methods based on KV cache pruning serve as efficient model serving strategies \cite{fu2024lazyllm,wan2024look}. 
\section{Linear Video Tokenizer for Video-LLMs}
\label{sec:method}

In Sec.~\ref{sec:overview}, we present the overall architecture of a video-LLM based on Linear Video Tokenizer (LinVT). Following this, Sec.~\ref{sec:linvt} introduces the two primary modules of LinVT: Spatio-Temporal Visual Token Refiner (SVR) described in Sec.~\ref{sec:method_1}, and Text-conditioned Token Aggregation (TTA) detailed in Sec.~\ref{sec:method_2}. Finally, we describe the training recipe for the video-LLM in Sec.~\ref{sec:training}.

\begin{figure*}
    \centering
    \includegraphics[width=1.0\linewidth]{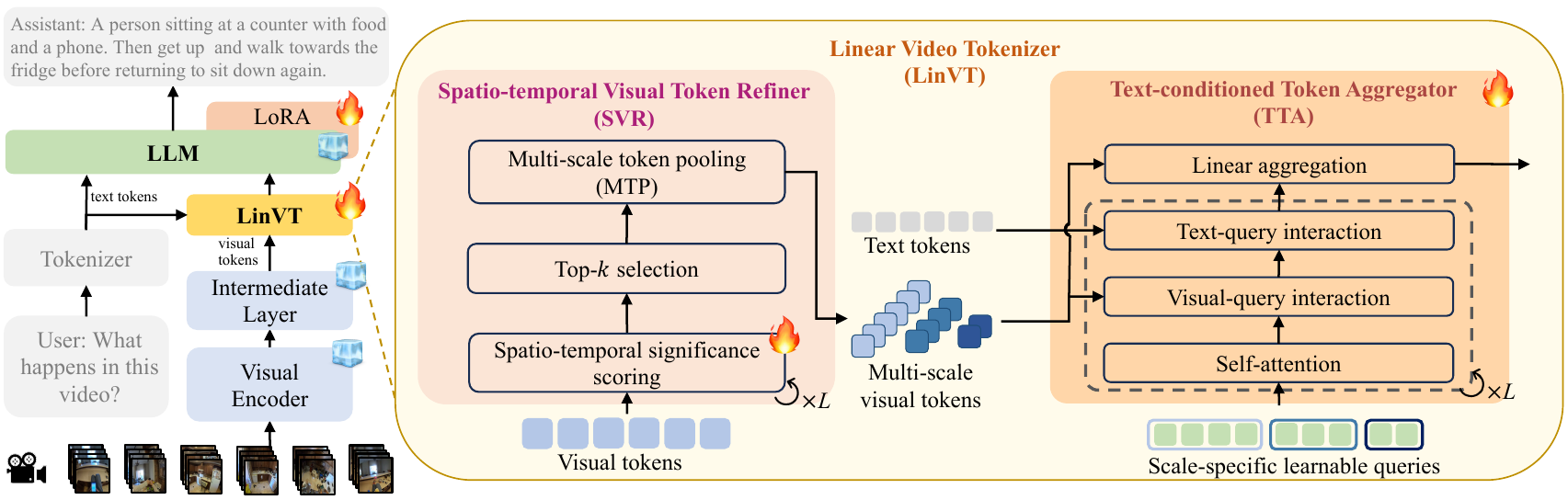}
    \caption{
    The framework of the LinVT-based video-LLM. The LinVT module takes visual tokens corresponding to individual frames of a video along with the user instruction as input and then generates compact and fixed-size visual tokens. 
    By using LinVT, an image-LLM can be easily converted to a video-LLM. 
    Firstly, the visual tokens undergo the spatio-temporal visual token refiner \emph{SVR} (Sec.~\ref{sec:method_1}) which produces multi-scale visual tokens.
    The multi-scale visual tokens are then fed to the text-conditioned token aggregator \emph{TTA} (Sec.~\ref{sec:method_2}).
    Finally, the LLM incorporates both the user instruction and the output visual tokens to provide a response for video understanding. The proposed LinVT operates linearly, enabling the preservation of knowledge from the image-LLM. 
    }
    \vspace{-2mm}
    \label{fig:details} 
\end{figure*}

\subsection{Video-LLM Architecture} \label{sec:overview}
The overall architecture of a video-LLM, derived from an image-LLM, is depicted in Fig.~\ref{fig:ffc_pipeline} (left part). A video-LLM typically comprises a text stream and a visual stream, which are subsequently processed by a large language model. 
In the visual stream, different image-LLMs may exhibit slightly varied structures. However, a common component is the visual encoder and the intermediate layers that project visual features into visual tokens. We do not elaborate on these intermediate layers, as they are not the primary focus of this work. Notably, the proposed LinVT is compatible with all the LLMs considered in this study.

\noindent\textbf{Information flow.}
Initially, the video frames are passed through the pre-trained visual encoder and some intermediate layers, generating visual tokens for each frame. LinVT then processes the visual tokens to represent the entire video in an informative and efficient way. 
The text tokenizer in the text stream transforms the input question into textual tokens. 
Finally, by concatenating the visual tokens and textual tokens and feeding them into the LLM, the answer is obtained in text format. 
In particular, LinVT-based video-LLMs can still \textbf{process image input}, and no modification to the model is required.

\subsection{Linear Video Tokenizer (LinVT)}
\label{sec:linvt}
\paragraph{Design principles of LinVT.}
As discussed in Sec.~\ref{sec:intro}, we have two fundamental design principles for the video tokenizer: (1) \textbf{linearity} -- the outputs of the video tokenizer should be the linear combination of its input tokens (2) \textbf{representative information condensation} -- the output only contains informative visual tokens that are representative for the video.
This first principle allows the video-LLM to attain image-level visual understanding without the need for training on a large corpus of image data. The second principle is to enhance the efficiency of the video-LLM to process the video tokens while preserving the rich contents of the video and thus improving the performance of video understanding. 
To achieve this, we propose a plug-and-play module called Linear Video Tokenizer (LinVT), which can be easily integrated into existing image-LLMs~\cite{zhu2024comprehensive, deitke2024molmo, xue2024xgen, chen2024far, wang2024qwen2, gu2024infinitymmscalingmultimodalperformance} to empower them with the ability to complete video understanding tasks, as depicted in Fig.~\ref{fig:ffc_pipeline}.

LinVT consists of two sub-modules, namely, Spatio-temporal Visual token Refiner (SVR) and Text-conditioned Token Aggregation (TTA). 
The two modules strictly follow the linearity principle, \ie the output of each module is the linear combination of (part of) its input, thereby preserving the visual-language alignment learned in the image-LLM.
On the other hand, the top-$k$ token selection in SVR and the TTA itself are two crucial steps to reduce the number of visual tokens passed to the subsequent computationally-expensive LLM, achieving the information condensation principle. Notably, since the representative information are condensed into the output, LinVT is also particularly effective in handling longer videos.
We will elaborate on SVR and TTA next.

\subsubsection{Spatio-Temporal Visual Token Refiner 
(SVR)} \label{sec:method_1}
The SVR module is a critical component in enhancing video understanding capabilities in a pre-trained image-LLM. The primary purpose of the SVR module is to efficiently distill the vast amount of information present in video data into a concise set of visual tokens that maintain the essential spatial and temporal characteristics. 

SVR consists of three modules: the spatio-temporal significance scoring, the top-{$k$} selection, and the multi-scale token pooling.
Given the visual tokens \( \{V_1, V_2, \ldots, V_T\} \) output from the intermediate layers (where \( V_i \) denotes the representation of the \( i \)-th frame and $T$ is the length of videos), SVR produces a refined set of visual tokens that encompass multi-scale visual tokens.
The three modules in SVR are introduced in detail as follows.

\noindent\textbf{Spatio-temporal significance scoring.}
The first visual token refinement step is achieved by spatio-temporal attention, which enables the input visual features to assess their relevance in spatial (intra-frame) and temporal (inter-frame) dimensions. This involves computing significance scores for each token, which reflects their importance in the entire video context. 

Concretely, for the visual token \( V_i  \in \mathbb{R}^{H \times W \times C} \) of the $i$-frame, where \( H \), \( W \) and \( C \) represent the height, width and the number of channels, respectively. The spatial significance score for each frame is obtained by applying self-attention within the corresponding visual token. Subsequently, the final significance scores are computed by applying self-attention across the sequence of spatial significance scores for the frames, thereby incorporating temporal information.

\noindent\textbf{Score-based token selection.}
To condense the information in the video, we select the top $k$ visual tokens with the highest significance scores and disregard the remaining tokens. The selected tokens are expected to represent the most informative aspects of the video in terms of spatial configuration and temporal dynamics.

\noindent\textbf{Multi-scale token pooling.}
Multi-scale representations play a crucial role in various video understanding tasks, including temporal action localization~\cite{zeng2025unimd, shi2023tridet,shi2022react,li2024detal}, moment retrieval~\cite{mu2024snag, pan2023scanning} and highlight detection~\cite{lei2021detecting, gordeev2024saliency,lin2023univtg}. In order to incorporate a multi-scale mechanism into the video-LLM, we introduce Multi-scale Token Pooling (MTP) after the top $k$ token selection. 
MTP analyzes these remaining $k$ visual tokens at various resolutions to capture both short-term and long-term dependencies within the video. 
Specifically, we use the average pooling in a shifted-window manner across the token or temporal axis.

\begin{figure}
    \centering
    \includegraphics[width=1.0\linewidth]{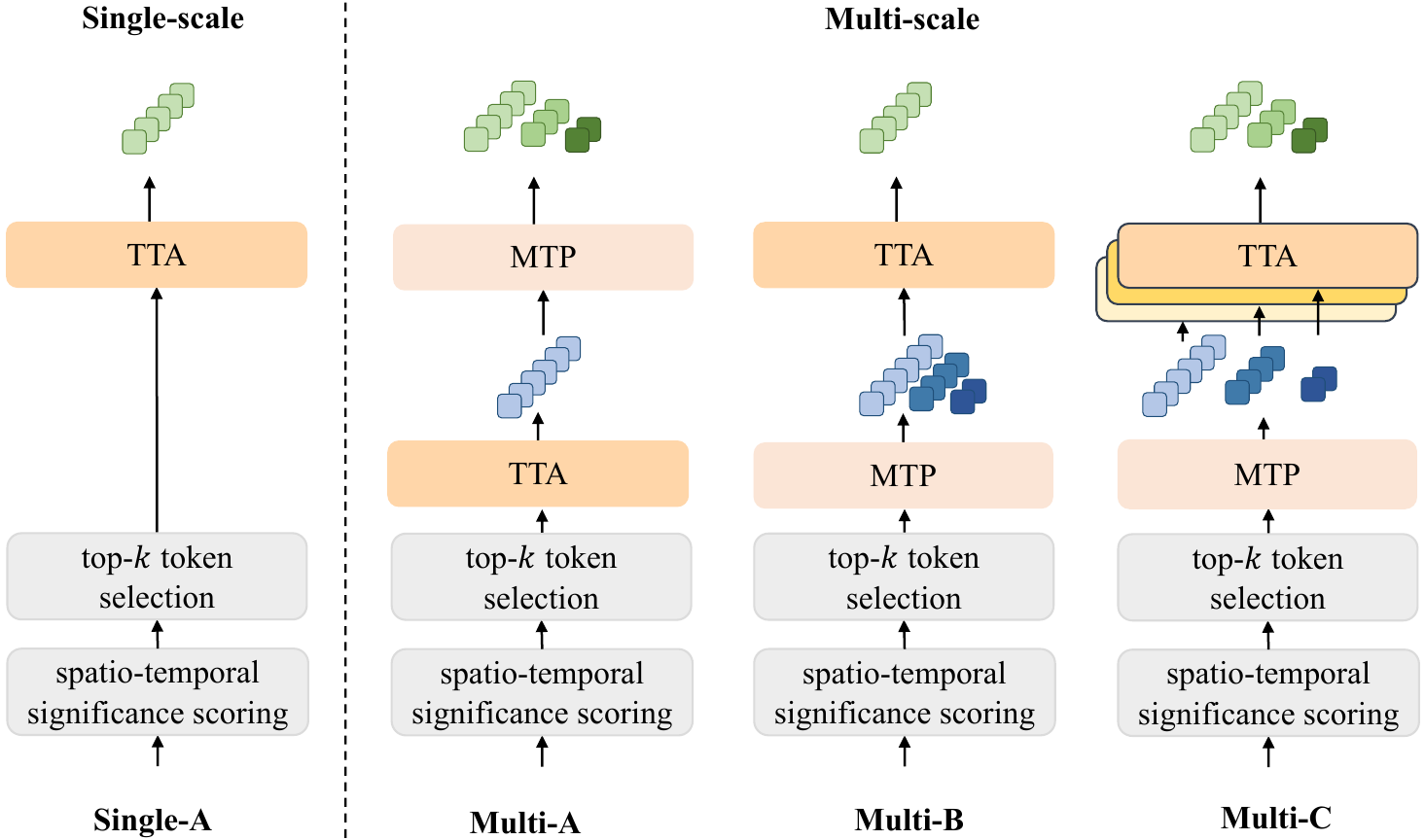}
    \caption{
    The left part represents single-scale token processing, while the right part contains three variants of multi-scale token processing in LinVT. For a fair comparison, all variants maintain the same output visual token size.
    }
    \label{fig:enter-label}
    \vspace{-2mm}
\end{figure}

In this work, we conduct a rigorous investigation of the designs of multi-scale token processing. Specifically, three variants are compared, as illustrated in Fig.~\ref{fig:enter-label} (right part).
\textbf{Multi-A}: MTP is placed after TTA. In this case, pooling is applied on the video tokens generated by TTA (see details of TTA in Sec.~\ref{sec:method_2}).
\textbf{Multi-B} and \textbf{Multi-C}: MTP is placed before TTA, and MTP operates on the top-$k$ selected visual tokens. 
The only difference between these two designs is whether TTA is applied on all scales at once (\ie Multi-B) or each scale separately (\ie Multi-C). 
By comparing these three designs experimentally (see Sec.~\ref{sec:ablation}), \textbf{Multi-C is the chosen design in our LinVT.}
The output of MTP is a set of multi-scale video tokens
\(MT_{v} = \{T_{v}^{1}, T_{v}^{2}, \ldots, T_{v}^l\}\), where $l$ is the number of scales. 

\noindent\textbf{Discussion.} The SVR module distinguishes itself from previous methods (such as TimeChat~\cite{ren2024timechat}, ATP~\cite{buch2022revisiting} and Non-local~\cite{wang2018non}) in two key aspects. First, the approach of selecting top-$k$ visual tokens are based on significance scores, which is a more efficient selection process compared to traditional methods that rely on fixed heuristics or exhaustive feature extraction. (2) While some existing methods consider temporal information, MTP offers a more comprehensive analysis by capturing dependencies at various temporal resolutions. This allows for better adaptation to various video challenges, such as rapid movements, motion blur, long video, \etc.

\subsubsection{Text-conditioned Token Aggregation (TTA)}\label{sec:method_2}
The TTA module is designed to refine and integrate multi-scale visual tokens with textual information, thereby enhancing the model's capacity to understand and interpret video content within a multimodal context. To achieve this, scale-specific queries are utilized to aggregate multi-scale visual tokens, which not only reduce the number of visual tokens but also integrate and enhance the semantic information at a higher level. Furthermore, we have developed a cross-modal decoder to combine image and text modal features. Each group of cross-modal queries is processed through a self-attention layer, an image cross-attention layer (to integrate image features), and a text cross-attention layer (to integrate text features). 
Lastly, the processed cross-modal queries are fed into the linear aggregation layer to perform linear aggregation on the multi-scale visual tokens.
This approach injects textual information into the queries to achieve improved modal alignment.

\noindent\textbf{Scale-specific learnable queries.} 
In TTA, we utilize scale-specific learnable queries for dynamic information extraction from the visual tokens and text tokens. Notably, the learnable queries are designed to generate attention maps across the visual tokens, which takes place in the final linear aggregation layer.
Compared to previous methods~\cite{alayrac2022flamingo,li2023blip} that process all tokens equally, utilizing scale-specific queries is a novel approach that allows the model to focus on the information across different scales. The queries are learnable, which are represented as \(MQ = \{\mathbf{Q}^{1}, \mathbf{Q}^{2}, \ldots, \mathbf{Q}^{l}\}\), where $l$ is the number of pre-defined scales. The self-attention mechanism operates exclusively within the same scales, without any attention across different scales.


\noindent\textbf{Visual-query interaction.}
After the self-attention process, the temporal attention queries and visual tokens from the SVR module engage in cross-attention. Note that the queries in each scale only interact with the visual tokens of the corresponding scale. This interaction helps to refines these visual tokens by focusing on specific aspects of the visual content that are relevant to the task. 

\noindent\textbf{Text-query interaction.}
Cross-attention with textual instruction serves as context attention to aggregate text-related visual features and enables more attention to visual tokens with high-response scores to input instruction. As a result, the most crucial visual cues related to user requirements are effectively stored in the output tokens of TTA.

\noindent\textbf{Linear aggregation layer.} The linear aggregation layer begins by calculating attention maps using the scale-specific queries $Q_{v,t}$ (after $L$ iterations of self-attention, visual-query and text-query interactions), in conjunction with the multi-scale visual tokens ($T_{v}$). These attention maps are subsequently used to aggregate the multi-scale visual tokens into fixed-length tokens. In other words, the output tokens of TTA represent various configurations of the weighted sum of multi-scale visual tokens, thereby enforcing linearity:
\begin{equation*}
\begin{aligned}
\Phi_{Lin}(T_{v}, Q_{v,t}) & = W_{attn} \cdot T_{v} \\
& = softmax(W_q \cdot Q_{v,t}~,~W_k \cdot T_{v}) \cdot T_{v},
\end{aligned}
\end{equation*}
where $\Phi_{Lin}$  is a linear function that computes different weighted sum of the visual tokens $T_{v}$, and $W_{attn}$ is the attention map derived from $T_{v}$ and ${Q_{v,t}}$. In contrast to vanilla cross-attention, $\Phi_{Lin}$ does not include residual connections and directly utilizes input values  $T_{v}$ without any transformations.

\noindent\textbf{Discussion.} The design of our TTA module differs from existing methods like Q-Former~\cite{instructblip} and Perceiver~\cite{alayrac2022flamingo} in several ways. First, the output of TTA is a linear combination of its input.
Second, TTA takes a more targeted approach in terms of multi-scale aggregation and text-driven visual token extraction, particularly in the context of video understanding. Our design places greater emphasis on handling temporal information and fusing multi-modal information. Moreover, while Perceiver also utilizes cross-attention mechanisms, the TTA module specifically adapts the query structure to accommodate temporal multi-scale characteristics. This further enhances the role of text in visual token compression, enabling TTA to effectively integrate temporal and semantic information in video understanding tasks.

\subsection{Training Recipe}
\label{sec:training}

In this work, we directly adapt the weights of the existing image-LLM, including the vision encoder, intermediate layer, and LLM. Then we utilize the LinVT module with randomly initialized parameters. Subsequently, it undergoes a two-stage training phase comprising video-language alignment training and video instruction tuning.

\noindent\textbf{Video-language alignment training.} 
 In this stage, we focus on enabling the model to comprehend video-text inputs through a video-language alignment dataset, which consist of 2.9 million video-text pairs. During this stage, the LLM model, vision encoder, and intermediate layer remain frozen, only the LinVT are learnable. Training is conduct under contrastive loss and cross-entropy loss.

\noindent\textbf{Video instruction tuning.} 
In the second stage, we perform video instruction tuning to enhance the model's capabilities in generating responses that align more closely with human preferences. During the training process, we incorporate a learnable LoRA~\cite{hu2021lora} into the LLM, while keeping the LinVT module learnable and freezing the other components. The training is conducted using the next-token prediction loss.
\section{Experiments}
\label{sec:experiment}

In this section, we present the experimental setup, conduct an ablation study of each component of LinVT (in Sec.~\ref{sec:ablation}), and provide comparisons with state-of-the-art video LLMs on general video understanding benchmarks (in Sec.~\ref{sec:gener_t}) and long-form video benchmarks (in Sec.~\ref{sec:long_t}). Implementation details and additional results can be found in the supplementary material.

\noindent\textbf{Baselines.} The proposed LinVT module has been incorporated into six recent multi-modal LLMs, namely
Aquila~\cite{gu2024infinitymmscalingmultimodalperformance}, Blip-3~\cite{xue2024xgen}, InternVL2~\cite{chen2024far}, Mipha~\cite{zhu2024comprehensive}, Molmo~\cite{deitke2024molmo} and Qwen2-VL~\cite{wang2024qwen2}.
It is notable that Blip-3, Mipha and Molmo are three image-LLMs that have not been trained on video data, whereas Aquila and InternVL2 have been primarily trained on image data and a limited amount of video data. Consequently, our experiments involve two types of baseline models: pure image-LLMs (\ie  Blip-3, Mipha and Molmo) and mixed image-LLMs which have been trained using image and video data (\ie Aquila, InternVL2 and Qwen2-VL).

\begin{table}[t!]
\centering
\resizebox{0.65\columnwidth}{!}{
\begin{tabular}{l|c}
\hline
{Datasets} & {Number of samples} \\
\hline
\multicolumn{2}{c}{\textit{Stage 1: video-language alignment training}} \\
\hline
WebVid-2M~\cite{bain2021frozen} & 2M \\
ShareGPTVideo~\cite{zhang2024direct} & 900k \\
\hline
\multicolumn{2}{c}{\textit{Stage2: video instruction tuning}} \\
\hline
VideoChatGPT~\cite{maaz2023video} & 90k \\
VideoChat2~\cite{li2024mvbench} & 230k \\
ActivityNet-QA~\cite{yu2019activitynet} & 29k \\
MSVD-QA~\cite{xu2017video} & 30k \\
MSRVTT-QA~\cite{xu2017video} & 109k \\
TGIF-QA~\cite{jang2017tgif} & 71k \\
NExT-QA~\cite{xiao2021next} & 34k \\
\hline
\end{tabular}
}
\caption{Details about video-language alignment training datasets and video instruction tuning datasets.}
\vspace{-1mm}
\label{tad:detail_dataset}
\end{table}

\noindent\textbf{Training data.} The datasets used for training are detailed in Tab.~\ref{tad:detail_dataset}. Note that we exclude the image data in VideoChat2~\cite{li2024mvbench}. Thus, only video data is used in both training stages.

\noindent\textbf{Evaluation benchmarks.} 
We report results on several video-based and image-based benchmarks. In particular, we evaluate the zero-shot performance on the open-ended video QA benchmark on MSVD-QA (MVD-QA) ~\cite{xu2017video}, MSRVTT-QA (MTT-QA)~\cite{xu2017video}, ActivityNet-QA (Act-QA)~\cite{yu2019activitynet} and TGIF-QA~\cite{jang2017tgif}. In addition, we include widely-adopted video understanding benchmarks, MVBench (MVB)~\cite{li2024mvbench}, MMBench-Video (MMB-V)~\cite{fang2024mmbench}, Video-MME (V-MME)~\cite{fu2024video},
NExT-QA (N-QA)~\cite{xiao2021next}, EgoSchema (EgoS)~\cite{mangalam2023egoschema}, VideoVista ~\cite{li2024videovista} and TempCompass (TempC)~\cite{liu2024tempcompass} for validation. The proposed model is also assessed on long video benchmarks like MLVU~\cite{zhou2024mlvu}, LongVideoBench (LongVB)~\cite{wu2024longvideobench} and the long-video subset of VideoMME~\cite{fu2024video}. Apart from the video domain, we perform regression evaluation on the following image benchmarks: {$\text{MMMU}_{\text{val}}$}~\cite{yue2024mmmu}, {$\text{DocVQA}_{\text{test}}$}~\cite{mathew2021docvqa}, {$\text{HallusionBench}_{\text{avg}}$}~\cite{guan2024hallusionbench}, {$\text{AI2D}_{\text{test}}$}~\cite{kembhavi2016diagram}, MMVet~\cite{yu2024mm}, {$\text{TextVQA}_{\text{val}}$}~\cite{singh2019towards}, POPE~\cite{li2023evaluating}, SEED-Image~\cite{li2024seed} and MME series (MME Perception and MME Cognition)~\cite{fu2023mme}. 
More experimental results on video and image benchmarks are provided in the supplementary material.

\subsection{Ablation Study} \label{sec:ablation}

\begin{table*}[t!]
\centering
\resizebox{\textwidth}{!}{%
\begin{tabular}{ccc|ccccccccccc}
\hline
Avg & SVR & TTA & {MVB} & {MMB-V} & {V-MME} & {LongVB} & {N-QA} & {EgoS} & {MVD-QA} & {MTT-QA} & {Act-QA} & {TGIF-QA}\\
\hline
\checkmark & & & 41.6 & 0.55 & 30.5 / 30.1 & 31.0 & 40.7 & 29.8 & 35.4 / 1.8 & 24.1 / 1.2 & 20.3 / 1.6 & 33.2 / 1.5 \\
& \checkmark & & 60.3 & 1.12 & 45.7 / 46.1 & 46.2 & 67.1 & 51.4 & 67.5 / 3.4 & 52.7 / 2.7 & 46.2 / 2.9 & 68.9 / 3.6\\
& & \checkmark & 58.7 & 1.03 & 45.4 / 45.7 & 44.8 & 65.1 & 50.2 & 64.1 / 3.2 & 50.3 / 2.4 & 41.6 / 2.6 & 66.9 / 3.2\\
& \checkmark & \checkmark & \textbf{62.5} & \textbf{1.22} & \textbf{48.4} / \textbf{49.2} & \textbf{49.7} & \textbf{71.1} & \textbf{55.9} & \textbf{71.2} / \textbf{3.8} & \textbf{55.3} / \textbf{3.0} & \textbf{47.5} / \textbf{3.1} & \textbf{71.1} / \textbf{3.9} \\
\hline
\end{tabular}%
}
\vspace{-1.5mm}
\caption{
Effectiveness of SVR and TTA evaluated under video benchmarks. In particular, \emph{Avg} denotes a naive baseline that extracts one token from each frame by simply average-pooling the visual tokens of each frame and is trained on the same video data as others.
The image-LLM in this experiment is Mipha-1.6B. }
\label{tab:sub_modules}
\end{table*}

\noindent\textbf{Roles of SVR and TTA.} 
As illustrated in \cref{tab:sub_modules}, the naive baseline, where the visual tokens of each frame are averaged into a single token, exhibits low performance across all video benchmarks. In contrast, the model equipped with SVR demonstrates significant metric improvements compared to the baseline. While the TTA module follows a similar trend, its improvements are less pronounced in comparison to SVR. These results underscore the respective importance of SVR and TTA. Notably, a substantial performance enhancement is achieved when both proposed components are utilized together.

\begin{table*}[]
\centering
\vspace{-1.5mm}
\resizebox{\textwidth}{!}{
\begin{tabular}{c|ccccc|cccc}
\hline
 \multirow{2}{*}{Methods}  & \multicolumn{5}{c|}{video benchmarks} & \multicolumn{4}{c}{image benchmarks} \\
& {MVB} & {MMB-V} & {V-MME} & {LongVB} & {MVD-QA} & $\text{MMMU}_{\text{val}}$ & $\text{AI2D}_{\text{test}}$ & $\text{TextVQA}_{\text{val}}$ & {MMVet} \\ 
\hline
Baseline & - & - & - & - & - & \textbf{30.5} & 59.6 & 70.5 & \textbf{23.5} \\
LinVT w/o alignment &  62.1 & 1.20 & {48.3 / 48.9} & 48.2 &  {70.8 / 3.7} &  24.1 & 46.6 & 61.3 & 17.4 \\ 
LinVT w/ alignment    &  \textbf{62.5} & \textbf{1.22} & \textbf{48.4 / 49.2} & \textbf{49.7} & \textbf{71.2 / 3.8} & \textbf{30.5} & \textbf{59.8} & \textbf{70.6} & \textbf{23.5} \\
\hline
\end{tabular}
}
\caption{Ablation studies on the 
importantce of maintaining the original vision-language alignment.
We conduct experiments on image and video benchmarks. Here, we utilize Mipha-1.6B to compare different settings of vision-language alignment. \emph{Baseline} denotes the original image-LLM.
``LinVT w/o alignment'' refers to the image-LLM randomly initialized intermediate layer while ``LinVT w/ alignment'' preserved the original weights.   }
\vspace{-1.5mm}
\label{tab:ablation_align}
\end{table*}

\noindent\textbf{Importantce of maintaining the original vision-language alignment.}
We train the model solely on video data, with or without the original intermediate layer, and validate its performance on image and video understanding benchmarks.
As a way to break the vision-language alignment of the original image-LLM (\ie LinVT-Mipha in this case), we randomly initialize the vision-language projector in Mipha. This model is denoted by \emph{LinVT w/o alignment}. 
As shown in \cref{tab:ablation_align}, maintaining the original alignment between vision and language marginally enhances performance on all video benchmarks.
More importantly, on image benchmarks, LinVT without the original alignment fails to develop a nuanced understanding of images compared to the original image-LLM and experiences significant performance degradation across all datasets. In contrast, keeping the original intermediate layer and adding LinVT achieves almost equal performance with the baseline in all image benchmarks, and even slightly outperforms it in $\text{DocVQA}_{\text{test}}$ and $\text{TextVQA}_{\text{val}}$, indicating that the introduction of our module does not compromise the original image-level knowledge. 

\begin{table}[t!]
\centering
\vspace{-1.5mm}
\resizebox{0.48\textwidth}{!}{
\begin{tabular}{c|ccccc}
\hline
Methods & {MVB} & {MMB-V} & {V-MME} & {LongVB} & {MVD-QA} \\
\hline
{Single-A} & 60.9 & 1.12 & 46.1 / 45.3 & 45.9 & 67.6/3.4 \\
{Multi-A} & 59.9 & 1.06 & 45.5 / 45.3 & 43.6 & 67.8/3.5 \\
{Multi-B} & 61.7 & 1.16 & 46.9 / 47.6 & 47.6 & 69.3/3.7 \\
{Multi-C} & \textbf{62.5} & \textbf{1.22} & \textbf{48.4 / 49.2} & \textbf{49.7} & \textbf{71.2/3.8} \\
\hline
\end{tabular}
}
\vspace{-1.5mm}
\caption{
Ablation studies on three variants of multi-scale processing in LinVT, evaluated under video benchmarks. Mipha-1.6B is used here. Among these variants, \emph{Multi-C} is chosen as our final design in LinVT.
}
\label{tab:ablation_multi}
\end{table}

\noindent\textbf{Multi-scale processing in SVR and TTA.}
\cref{sec:method_1} introduces three variants for applying multi-scale techniques in LinVT. \cref{tab:ablation_multi} shows the examination of these methods. Among them, the multi-A method has a negative effect on video understanding compared to the single-scale processing (except for MVD-QA). This can be attributed to the fact that multi-A is not based on multi-scale temporal processing. The video-level vision tokens generated by TTA contain global representations, but the neighboring tokens are not strictly temporally correlated. 
The multi-B and multi-C methods achieve superior performance by integrating MTP before TTA, suggesting that separate multi-scale processing can more fully capture both long-term and short-term temporal information of visual tokens. Furthermore, the multi-C method shows the best performance across the video benchmark with scale-specific queries in TTA.

\begin{table}[t!]
\centering
\vspace{-1.5mm}
\resizebox{0.48\textwidth}{!}{
\begin{tabular}{c|ccccc}
\hline
Text Conditions & {MVB} & {MMB-V} & {V-MME} & {LongVB} & {MVD-QA} \\
\hline
& 61.8 & 1.19 & 47.6 / 48.9 & 49.1 & 70.1 / 3.6 \\
\checkmark & \textbf{62.5} & \textbf{1.22} & \textbf{48.4 / 49.2} & \textbf{49.7} & \textbf{71.2 / 3.8} \\
\hline
\end{tabular}
}
\caption{Ablation studies on the effectiveness of text conditions in TTA. The image-LLM used in this experiment is Mipha-1.6B. Interaction with the text conditions boosts performance on several video benchmarks.}
\vspace{-1.5mm}
\label{tab:ablation_text}
\end{table}

\noindent\textbf{Role of text conditions in TTA.} As shown in \cref{tab:ablation_text}, the interaction between text conditions and queries in TTA enables queries to assimilate additional information from text prompts. This interaction leads to enhanced performance on all video benchmarks when compared to the baseline model that does not incorporate text conditions.

\begin{table}[t!]
\centering
\vspace{-1.5mm}
\resizebox{0.48\textwidth}{!}{%
\begin{tabular}{c|c|ccccc}
\hline
Methods & Size & {MVB} & {MMB-V} & {V-MME} & {LongVB} & {MVD-QA} \\
\hline
{InternVL2} & \multirow{2}{*}{1B} &  57.9 & 0.93 & 42.6 / 44.7 & 46.2 & 51.4 / 2.8 \\
{LinVT-InternVL2} &  & \textbf{60.2} & \textbf{1.19} & \textbf{49.9 / 51.8} & \textbf{51.0} & \textbf{73.3 / 3.9} \\
\hline
{Aquila} & \multirow{2}{*}{2B} & 58.1 & 1.12 & 47.1 / 48.4 & 48.5 & 49.2 / 2.1 \\
{LinVT-Aquila} &  & \textbf{62.2} & \textbf{1.26} & \textbf{53.4 / 55.3} & \textbf{53.5} & \textbf{74.6 / 4.1} \\
\hline
{Qwen2-VL} & \multirow{2}{*}{7B} & 67.0 & 1.29 & 63.3 / 69.0 & 53.1 & 75.3 / 4.2 \\
{LinVT-Qwen2-VL} &  & \textbf{69.3} & \textbf{1.62} & \textbf{66.4 / 67.6} & \textbf{57.2} & \textbf{80.2 / 4.4} \\
 \hline
\end{tabular}
}
\caption{
Ablation studies on the effectiveness of LinVT in video-compatible visual LLMs. 
}
\label{tab:aquila_qwen2}
\end{table}

\noindent\textbf{Role of LinVT in video-compatible image-LLMs.} 
As illustrated in \cref{tab:aquila_qwen2}, for image-LLMs that include some video data in the training and can handle videos in the first place, namely InternVL2 and Aquila, the integration of LinVT can also significantly enhance their performance in video benchmarks.

\subsection{Comparison with Sate-of-The-Art} \label{sec:gener_t}

\begin{table}[t!]
\centering
\centering
\renewcommand{\arraystretch}{1.2}
\vspace{-4mm}
\resizebox{0.48\textwidth}{!}{
\begin{tabular}{l|c|cccc}
\hline
{Method} & {Size} & MVD-QA & MTT-QA & Act-QA & TGIF-QA \\
\hline
VideoChat~\cite{li2023videochat} & 7B & {56.3 / 2.8} & {45.0 / 2.5} & {- / 2.2} & {34.4 / 2.3} \\
Video-LLaMA~\cite{zhang2023video} & 7B & {51.6 / 2.5} & {29.6 / 1.8} & {12.4 / 1.1} & {- / -} \\
Video-LLaMA2~\cite{cheng2024videollama} & 7B & {71.7 / 3.9} & {- / -} & {49.9 / 3.3} & {- / -} \\
Video-ChatGPT~\cite{maaz2023video} & 7B & {64.9 / 3.3} & {49.3 / 2.8} & {34.2 / 2.8} & {51.4 / 3.0} \\
Chat-UniVi~\cite{jin2024chat} & 7B & {69.3 / 3.7} & {55.0 / 3.1} & {46.1 / 3.3} & {69.0 / 3.8} \\
LLaMA-VID~\cite{li2025llama} & 7B & {69.7 / 3.7} & {57.7 / 3.2} & {47.4 / 3.3} & {- / -} \\
Video-LLaVA~\cite{lin2023video} & 7B & {71.8 / 3.9} & {59.2 / 3.5} & {45.3 / 3.3} & {70.0 / 4.0} \\
MiniGPT4-Video~\cite{ataallah2024minigpt4} & 7B & {73.9 / 4.1} & {59.7 / 3.3} & {46.3 / 3.4} & {72.2 / 4.1} \\
PLLaVA~\cite{xu2024pllava} & 7B & {76.6 / 4.1} & {62.0 / 3.5} & {56.3 / 3.5} & {77.5 / 4.1} \\
SlowFast-LLaVA~\cite{xu2024slowfast} & 7B & {79.1 / 4.1} & {65.8 / 3.6} & {56.3 / 3.4} & {78.7 / 4.2} \\
Tarsier~\cite{wang2024tarsier} & 7B & {77.0 / 4.1} & {62.0 / 3.5} & {59.5 / 3.6} & {79.2 / 4.2} \\
BLIP-3-Video~\cite{ryoo2024xgen} & 4B & {77.7 / 4.2} & {60.0 / 3.6} & {55.7 / 3.5} & {76.5 / 4.3} \\
\hline
\rowcolor[HTML]{EFEFEF}
LinVT-Mipha & 1.6B & {71.2 / 3.8} & {55.3 / 3.0} & {47.5 / 3.1} & {71.1 / 3.9}\\
\rowcolor[HTML]{EFEFEF} 
LinVT-Aquila & 2B & 74.6 / 4.1 & 58.4 / 3.2 & 51.1 / 3.3 & 73.6 / 4.0 \\
\rowcolor[HTML]{EFEFEF} 
LinVT-BLIP-3 & 4B & 79.1 / 4.4 & 61.5 / 3.9 & 58.9 / 3.6 & 78.7 / 4.3 \\
\rowcolor[HTML]{EFEFEF} 
LinVT-Molmo & 7B & 78.1 / 4.3 & 60.3 / 3.7 & {59.6} / 3.7 & {79.3} / 4.2 \\
\rowcolor[HTML]{EFEFEF} 
LinVT-Qwen2-VL & 7B & \textbf{80.2} / 4.4 & \textbf{66.2} / 4.0 & \textbf{60.1} / 3.6 & \textbf{81.3} / 4.3 \\
\rowcolor[HTML]{EFEFEF} 
LinVT-InternVL2 & 8B &  \closeruline{79.3 }/ 4.4 & \closeruline{63.1} / 4.0 & \closeruline{59.7} / 3.7 & \closeruline{79.4} / 4.3 \\
\hline
\end{tabular}%
}
\vspace{-0.5em}
\caption{
Comparison with the state-of-the-art on four zero-shot open-ended video QA datasets. 
The best results are in \textbf{bold} and second best \ul{underlined}.
}
\label{tab:res_openend}
\end{table}

\noindent\textbf{Results on zero-shot open-ended video QA benchmark.}
In ~\cref{tab:res_openend}, we present a comparative evaluation of LinVT-based models against various State-of-The-Art (SoTA) methods across four zero-shot open-ended video QA benchmarks. It is evident that the LinVT-Qwen2-VL achieves the highest accuracy on all QA datasets, with 80.2\% on MVD-QA, 66.2\% on MTT-QA, 60.1\% on ActivityNet-QA and 81.3\% on TGIF-QA. Notably, The LinVT-Aquila model, despite having only 2B parameters, surpassed several 7B models. Specifically, it ranked fifth on the ActivityNet-QA benchmark, fifth on the TGIF-QA benchmark, and fifth on the MSVD-QA benchmark. Although training solely on video data, our models still achieve impressive results.

\noindent\textbf{Results on short and general video benchmarks.}
We also conduct experiments on six short and general video benchmarks, as shown in \cref{tab:general_res}. Our models obtain the best result among the counterparts with the same parameters. 
Moreover, our smallest model, LinVT-Mipha is still comparable to several 7B models such as Qwen-VL-Chat, LongVILA, Video-LLaVA, and VideoLLaMA2.

\begin{table}[t!]
\vspace{-4mm}
\centering
\renewcommand{\arraystretch}{1.1}
\resizebox{0.48\textwidth}{!}{%
\begin{tabular}{l|c|ccccccc}
\hline
Model & Size & MVB & MMB-V & EgoS & N-QA & VideoVista & TempC \\
\hline
VideoChatGPT~\cite{maaz2023video} & 7B & 32.7 & 0.93 & - & & 36.65 & 43.5 \\
{Video-LLaVA}~\cite{lin2023video} & 7B & 43.5 & 1.05 & 38.4 & - & 56.96 & 49.8 \\

LLaMA-VID~\cite{li2025llama} & 7B & 41.9 & - & 38.5 & - & 56.87 & 45.6 \\
PLLaVA~\cite{xu2024pllava} & 7B & 46.6 & 1.03 & 54.4 & - & 60.36 & - \\
VideoLLaMA2~\cite{cheng2024videollama} & 7B & 54.6 & - & 51.7 & - & 60.47 & - \\
LLaVA-NeXT-Video~\cite{zhang2024llavanextvideo} & 7B & 53.1 & - & 43.9 & 70.2 & 56.66 & 53.7 \\
TimeChat~\cite{ren2024timechat} & 7B & - & - & 33.0 & - & - & 50.6 \\
VideoChat2-HD~\cite{li2024mvbench} & 7B & 62.3 & - & 55.8 & 79.5 & 61.58 & - \\
Qwen2-VL~\cite{wang2024qwen2} & 7B & 67.0 & - & 66.7 & - & - & - \\
ShareGPT4Video~\cite{chen2024sharegpt4video} & 8B & 51.2 & 1.05 & - & - & 53.58 & 61.5 \\
VILA-1.5~\cite{lin2024vila} & 8B & - & - & 50.4 & - & 64.18 & 58.8 \\
Kangaroo~\cite{liu2024kangaroo} & 8B & 61.1 & 1.44 & 62.7 & - & 69.50 & 62.5 \\
Video-CCAM~\cite{fei2024video} & 9B & {64.6} & - & - & - & 69.00 & - \\
\hline
\rowcolor[HTML]{EFEFEF} 
LinVT-Mipha & 1.6B & 60.5 & 1.06 & 48.4 & 71.1 & 55.62 & 45.2 \\
\rowcolor[HTML]{EFEFEF} 
LinVT-Aquila & 2B & 59.2 & 1.12 & 57.9 & 76.6 & 60.13 & 52.2 \\
\rowcolor[HTML]{EFEFEF} 
LinVT-BLIP-3 & 4B & 63.7 & 1.53 & 62.9 & 80.1 & 68.62 & 59.6 \\
\rowcolor[HTML]{EFEFEF} 
LinVT-Molmo & 7B & 63.1 & {1.54} & {63.1} & 81.4 & {77.41} & {63.5} \\
\rowcolor[HTML]{EFEFEF} 
LinVT-Qwen2-VL & 7B & \textbf{69.3} & \textbf{1.62} & \textbf{69.5} & \textbf{85.5} & \textbf{79.67} & \textbf{65.8} \\
\rowcolor[HTML]{EFEFEF} 
LinVT-InternVL2 & 8B & \closeruline{67.7} & \closeruline{1.57} & \closeruline{63.9} & \closeruline{81.9} & \closeruline{78.64} & \closeruline{64.4} \\
\hline
\end{tabular}%
}

\caption{
Comparison with the state-of-the-art on six short and general video benchmarks. 
The best results are in \textbf{bold} and second best \ul{underlined}.
}
\label{tab:general_res}
\vspace{-2mm}
\end{table}

\begin{table}[t!]
\centering
\renewcommand{\arraystretch}{1.1}
\resizebox{0.48\textwidth}{!}{
\begin{tabular}{l|c|ccccc}
\hline
\multirow{2}{*}{Model} & \multirow{2}{*}{Size} & \multirow{2}{*}{MLVU} & \multirow{2}{*}{LongVB} & \multicolumn{2}{c}{VideoMME-Long} \\
 & & & & w/o subs & w subs \\
\hline
{GPT4-V}~\cite{openai20234v} & - & 49.2 & 60.7 & 53.5 & 56.9 \\
{GPT4-O}~\cite{zhu2024openai} & - & 64.6 & 66.7 & 65.3 & 72.1 \\
Gemini 1.5 Pro~\cite{team2024gemini} & - & - & 64.4 & 67.4 & 77.4 \\
\hline
VideoLLaMA2~\cite{cheng2024videollama} & 7B & 48.5 & - & 42.1 & 43.8 \\
LongVA~\cite{zhang2024long} & 7B & 56.3 & - & 46.2 & 47.6 \\
Qwen-VL-Chat~\cite{bai2023qwen} & 7B & - & - & 37.8 & 37.9\\
Kangaroo~\cite{liu2024kangaroo} & 8B & 61.0 & 54.8 & 46.6 & 49.3 \\
LongVILA~\cite{xue2024longvila} & 8B & - & - & 39.7 & - \\
VideoCCAM~\cite{fei2024video} & 14B & 63.1 & - & 46.7 & 49.9 \\
PLLaVA~\cite{xu2024pllava} & 34B & - & 53.5 & - & - \\
ShareGPT4Video~\cite{chen2024sharegpt4video} & 8B & 46.4 & 41.8 & 35.0 & 37.9 \\
{Video-LLaVA}~\cite{lin2023video} & 7B & 47.3 & 37.6 & 39.9 & 41.6 \\
LLaVA-OneVision~\cite{li2024llava} & 7B & 64.7 & - & - & - \\
Oryx~\cite{liu2024oryx} & 7B & 67.5 & 55.3 & 50.3 & 55.8 \\
\hline
\rowcolor[HTML]{EFEFEF} 
LinVT-Mipha & 1.6B & 56.2 & 49.7 & 44.5 & 46.1 \\
\rowcolor[HTML]{EFEFEF} 
LinVT-Aquila & 2B & 65.1 & 53.5 & 53.4 & 55.3\\
\rowcolor[HTML]{EFEFEF} 
LinVT-BLIP-3 & 4B & {67.9} & 56.6 & {58.3} & {62.4} \\
\rowcolor[HTML]{EFEFEF} 
LinVT-Molmo & 7B & 67.7 & {56.8} & 57.6 & 61.2 \\
\rowcolor[HTML]{EFEFEF} 
LinVT-Qwen2-VL & 7B & \textbf{68.9} & \closeruline{57.2} & \textbf{63.1} & \closeruline{63.3} \\
\rowcolor[HTML]{EFEFEF} 
LinVT-InternVL2 & 8B & \closeruline{68.3} & \textbf{57.4} & \closeruline{60.2} & \textbf{63.5} \\
\hline
\end{tabular}
}
\caption{
Comparison with the state-of-the-art on three long video benchmarks. The best results of open-source models are in \textbf{bold} and second best \ul{underlined}.
}
\vspace{-2mm}
\label{tab:long_video}
\end{table}

\noindent\textbf{Results on long-form video benchmarks} \label{sec:long_t}
As shown in \cref{tab:long_video}, LinVT-InternVL2, LinVT-Qwen2-VL, LinVT-Molmo and LinVT-BLIP-3 have been validated as achieving the top four results among open-source models. LinVT-Qwen2-VL significantly surpasses others, with an improvement margin of +12.8\% in VideoMME-Long (without subtitles), while LinVT-InternVL2 achieves a 7.7\% increase in VideoMME-Long (with subtitles). Even with a smaller size of 2B parameters, LinVT-Aquila demonstrates competitive capabilities in handling long videos. Although it ranks behind Kangaroo and Oryx in LongVideoBench, its performance is comparable to that of the 34B model \emph{PLLaVA}. Overall, LinVT enhances models with robust capabilities for processing long videos, highlighting its effectiveness in visual condensation.

\begin{figure}
    \centering
    \includegraphics[width=1.0\linewidth]{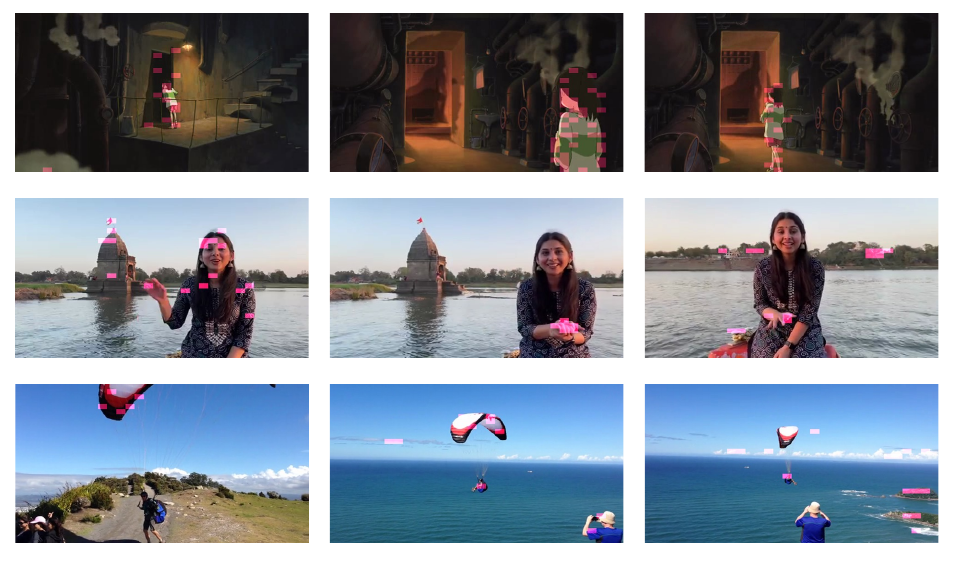}
    \vspace{-5.0mm}
    \caption{Visualization of the patches corresponding to the selected tokens in video frames. Each row corresponds to a video.
    The selection is achieved by the spatio-temporal significance scoring and top-$k$ selection. The \textcolor{red}{\textbf{red patches}} in the image represent the selected tokens. This token scoring and selection mechanism directs the attention of the model towards the most prominent objects, actions, or scenes within the video. 
    LinVT-InternVL2-8B is used for this visualization.
    }
\vspace{-3mm}
\label{fig:selector}
\end{figure}

\subsection{Qualitative Analyses} 
\noindent\textbf{Visualization of selected tokens.}
In \cref{fig:selector}, we visualize the corresponding patches of the selected tokens. The spatio-temporal significance scoring and selection mechanism guides the model to focus on regions that most significantly contribute to the task objective. For instance, in the first row of \cref{fig:selector}, the selected tokens are concentrated on the individuals present. In the second row, the focus shifts from the scene to the foreground of the frame, and then to the hands of the individuals in the image. This approach allows for the elimination of redundant or irrelevant information from the video while retaining the most critical tokens, thereby enhancing both performance and computational efficiency.
\section{Conclusion}
This study proposes a module called LinVT, which enhances the video understanding capability by integrating it into existing image-level large language models. The LinVT module efficiently processes video data, particularly in long video tasks, while maintaining the original image understanding capability. By using spatio-temporal visual token refiners and text-conditioned token aggregators, LinVT effectively compresses and integrates multimodal information, enhancing the representation capability of video tokens. Experimental results demonstrate that LinVT achieves competitive performance in multiple video question-answering tasks and image question-answering benchmarks, validating its effectiveness in multimodal video understanding. Future research can further optimize the structure and performance of LinVT to address more complex video understanding tasks.
\label{sec:conclusion}
{
    \small
    \bibliographystyle{ieeenat_fullname}
    \bibliography{main}
}

\clearpage
\setcounter{page}{1}

\twocolumn[{%
\renewcommand\twocolumn[1][]{#1}%
\maketitlesupplementary
\begin{center}
\begin{table}[H]
\centering
\resizebox{\textwidth}{!}{
\begin{tabular}{c|cccccc|cccccc}
\hline
\multirow{2}{*}{Methods} & \multicolumn{6}{c|}{video benchmarks} & \multicolumn{6}{c}{image benchmarks} \\
 & {LongVB} & {N-QA} & {EgoS} & {MTT-QA} &{Act-QA} & {TGIF-QA} & $\text{DocVQA}_{\text{test}}$ & $\text{HallBench}_{\text{avg}}$ & {POPE} & {SEED-Image} & {MME Perception} & {MME Cognition} \\ 
\hline
Original image-LLM & - & - & - & - & - & - & 75.3 & \textbf{28.2} & 86.9 & \textbf{56.1} & 1230.1 & \textbf{247.9} \\
LinVT w/o alignment &  \textbf{49.6} & 70.9 & 55.4 & {55.1 / 2.9}  &  {47.1 / 3.0} & {70.6 / 3.8} & 65.4 & 20.2 & 78.2 & 39.2 & 1065.7 & 216.8 \\ 
LinVT w/ alignment    &  \textbf{49.6} & \textbf{71.1} & \textbf{55.9} & \textbf{55.3 / 3.0} &  \textbf{47.5 / 3.1} & \textbf{71.1 / 3.9} & \textbf{75.4} & 28.1 & \textbf{87.1} & 55.8 & \textbf{1235.5} & 247.7 \\
\hline
\end{tabular}
}
\end{table}
\vspace{-3.5mm}
\captionof{table}{Ablation studies on more video and image benchmarks about the capability to maintain original image knowledge. Here, we employ Mipha-1.6B as the image-LLM. The baseline simply averages the visual tokens from different frames.~\emph{LinVT w/o alignment} refers to the image-LLM randomly initialized intermediate layer while \emph{LinVT w/ alignment} preserved the original weights.}
\label{tab:suppl_random_projector_video}
\end{center}
}]

\section{Implementation details} 
Unless specified in the ablation study, SVR consists of $L=4$ spatio-temporal layers, while the TTA module incorporates multi-scale processing with $l=3$ different scales (\ie 64, 32, and 16). The number of multi-heads in the TTA module is set to 8. Prior to inputting the frame-level visual tokens into the LinVT module, a temporal boundary detection model called AutoShot~\cite{zhuautoshot} is used to segment videos into clips. 
Noteworthy, LinVT is lightweight and its number of parameters is 267M, which is negligible compared to the original billion-size image-LLMs.

\section{More Experiments for Ablation Study}
\label{sec:suppl_ablation}

\paragraph{Importantce of maintaining the original vision-language alignment.}
More validation results are presented in \cref{tab:suppl_random_projector_video}. The previous conclusion in \cref{tab:ablation_align} holds in the additional benchmarks.

\paragraph{Role of LinVT in video-compatible image-LLMs.} 
As illustrated in \cref{tab:suppl_internvl_aquila}, for image-LLMs that include some video data in the training and can handle videos in the first place, namely InternVL2, Aquila, and Qwen2-VL, the integration of LinVT can also significantly enhance their performance in video benchmarks.

\paragraph{Multi-scale processing in SVR and TTA.}
\cref{tab:suppl_ablation_multi} presents a comprehensive evaluation of different designs of multi-scale processing. The subsequent results further validate our final selection, the multi-C method, which demonstrates superior performance across the video benchmark.

\paragraph{Role of text conditions in TTA.} As demonstrated in \cref{tab:suppl_textcondition}, the interaction between text conditions and queries consistently results in enhanced performance across additional video benchmarks, beyond the results shown in \cref{tab:ablation_text}, when compared to the baseline model that does not incorporate text conditions.

\paragraph{Design choices of LinVT.} 
We conduct ablation experiments to evaluate the design choices of LinVT structures, focusing on specific parameters: the number of layers in the block layer of SVR and TTA, the number of scale-specific learnable queries, and the number of selected visual tokens in the top $k$ selection in SVR. As shown in \cref{tab:suppl_hyperparmeter}, doubling the number of parameters does not lead to a significant improvement in performance metrics, but it does increase the computational cost. Conversely, a substantial performance degradation is observed when the parameters are reduced by half. These findings suggest that our current settings achieve an optimal balance between performance and computational efficiency.

\begin{table*}[h]
\resizebox{\textwidth}{!}{%
\begin{tabular}{c|c|cccccccccc}
\hline
Methods & Size & {MVB} & {MMB-V} & {V-MME} & {LongVB} & {N-QA} & {EgoS} & {MVD-QA} & {MTT-QA} & {Act-QA} & {TGIF-QA}\\
\hline
${\text{InternVL2}}^{*}$ & 1B & 57.9 & 0.98 & 42.6 / 44.7 & - & - & - & - & - & - & - \\
{InternVL2} & 1B &  57.9 & 0.93 & 42.6 / 44.7 & 46.2 & 67.6 & 49.3 & 51.4 / 2.8 & 40.7 / 2.6 & 38.6 / 3.0 & 49.3 / 3.1 \\
{LinVT-InternVL2} & 1B & \textbf{60.2} & \textbf{1.19} & \textbf{49.9 / 51.8} & \textbf{51.0} & \textbf{72.4} & \textbf{55.7} & \textbf{73.3 / 3.9} & \textbf{57.1 / 3.0} & \textbf{49.6 / 3.1} & \textbf{72.2 / 4.0}\\
\hline
${\text{Aquila}}^{*}$ & 2B & - & - & {- / 48.4} & - & - & - & - & - & - & - \\
{Aquila} & 2B & 58.1 & 1.12 & 47.1 / 48.4 & 48.5 & 67.1 & 51.4 & 49.2 / 2.1 & 35.5 / 2.4 & 29.7 / 2.6 & 42.5 / 2.8 \\
{LinVT-Aquila} & 2B & \textbf{62.2} & \textbf{1.26} & \textbf{53.4 / 55.3} & \textbf{53.5} & \textbf{76.6} & \textbf{57.9} & \textbf{74.6 / 4.1} & \textbf{58.4 / 3.2} & \textbf{51.1 / 3.3} & \textbf{73.6 / 4.0} \\
\hline
${\text{Qwen2-VL}}^{*}$ & 7B & 67.0 & - & 63.3 / 69.0 & - & - & 66.7 & - & - & - & - \\
{Qwen2-VL} & 7B & 67.0 & 1.29 & 63.3 / 69.0 & 53.1 & 79.4 & 66.7 & 75.3 / 4.2 & 59.1 / 3.4 & 51.9 / 3.3 & 74.4 / 4.0 \\
{LinVT-Qwen2-VL} & 7B & \textbf{69.3} & \textbf{1.62} & \textbf{66.4 / 67.6} & \textbf{57.2} & \textbf{85.5} & \textbf{69.5} & 
 \textbf{80.2 / 4.4} & \textbf{66.2 / 4.0} & \textbf{60.1 / 3.6} & \textbf{81.3 / 4.3} \\
 \hline
\end{tabular}
}
\caption{
Ablation studies on the effectiveness of LinVT in video-compatible image-LLMs. 
The results of the \emph{InternVL2}, \emph{Aquila} and \emph{Qwen2-VL} are obtained using their default video processing pipeline. * represents the accuracy reported in the relevant papers.
}
\label{tab:suppl_internvl_aquila}
\end{table*}
\begin{table*}[h!]
\resizebox{\textwidth}{!}{%
\begin{tabular}{ccc|ccccccccccc}
\hline
{\#layers} & {\#queries} & {$k$} & {MVB} & {MMB-V} & {V-MME} & {LongVB} & {N-QA} & {EgoS} & {MVD-QA} & {MTT-QA} & {Act-QA} & {TGIF-QA}\\
\hline
4 & {\{64, 32, 16\}} & 2048 & 62.5 & 1.22 & 48.4 / 49.2 & 49.7 & 71.1 & 55.9 & 71.2/3.8 & 55.3/3.0 & 47.5/3.1 & 71.1/3.9 \\
8 & {\{64, 32, 16\}} & 2048 & 62.7 & 1.23 & 48.9 / 49.5 & 50.6 & 72.3 & 56.7 & 72.0/4.0 & 55.7/3.1 & 47.9/3.2 & 71.7/4.0 \\
2 & {\{64, 32, 16\}} & 2048 & 58.1 & 1.05 & 44.6 / 44.8 & 45.1 & 65.2 & 50.1 & 65.4/3.2 & 49.8/2.4 & 43.3/2.7 & 64.7/3.4 \\
\hline
4 & {\{64, 32, 16\}} & 2048 & 62.5 & 1.22 & 48.4 / 49.2 & 49.7 & 71.1 & 55.9 & 71.2/3.8 & 55.3/3.0 & 47.5/3.1 & 71.1/3.9 \\
4 & {\{128, 64, 32\}} & 2048 & 62.8 & 1.23 & 48.8 / 49.9 & 51.4 & 71.6 & 56.7 & 71.5/3.9 & 55.4/3.0 & 47.6/3.1 & 72.6/4.0 \\
4 & {\{32, 16, 8\}} & 2048 & 59.6 & 1.09 & 45.1 / 44.7 & 45.2 & 68.6 & 52.5 & 70.1/3.7 & 53.7/2.9 & 46.5/2.9 & 68.7/3.6\\
\hline
4 & {\{64, 32, 16\}} & 2048 & 62.5 & 1.22 & 48.4 / 49.2 & 49.7 & 71.1 & 55.9 & 71.2/3.8 & 55.3/3.0 & 47.5/3.1 & 71.1/3.9 \\
4 & {\{64, 32, 16\}} & 4096 & 63.1 & 1.25 & 48.6 / 49.7 & 50.3 & 71.8 & 55.6 & 71.4/3.9 & 55.6/3.1 & 47.5/3.1 & 71.2/3.9\\
4 & {\{64, 32, 16\}} & 1024 & 61.1 & 1.02 & 46.5 / 47.1 & 46.2 & 68.3 & 52.4 & 67.7/3.4 & 51.1/2.5 & 42.2/2.6 & 68.3/3.6\\
\hline
\end{tabular}
}
\caption{Ablation studies of the design choice of LinVT structures. We examine the impact of several parameters within LinVT, including the number of layers in the blocks of SVR and TTA, the number of scale-specific learnable queries, and the value of $k$ in top-$k$ selection. For all metrics assessed, higher values are preferred.}
\label{tab:suppl_hyperparmeter}
\end{table*}

\begin{table*}[h!]
\centering
\begin{tabular}{c|cccccccccc}
\hline
Methods & {MVB} & {MMB-V} & {V-MME} & {LongVB} & {N-QA} & {EgoS} & {MVD-QA} & {MTT-QA} & {Act-QA} & {TGIF-QA}\\
\hline
{Single-A} & 60.9 & 1.12 & 46.1 / 45.3 & 45.9 & 67.3 & 51.8 & 67.6/3.4 & 52.5/2.6 & 46.2/2.9 & 68.6/3.5 \\
{Multi-A} & 59.9 & 1.06 & 45.5 / 45.3 & 43.6 & 65.5 & 52.1 & 67.8/3.5 & 52.5/2.6 & 44.3/2.8 & 68.3/3.7 \\
{Multi-B} & 61.7 & 1.16 & 46.9 / 47.6 & 47.6 & 70.8 & 53.5 & 69.3/3.7 & 53.1/2.8 & 46.4/3.0 & 69.8/3.8 \\
{Multi-C} & \textbf{62.5} & \textbf{1.22} & \textbf{48.4 / 49.2} & \textbf{49.7} & \textbf{71.1} & \textbf{55.9} & \textbf{71.2/3.8} & \textbf{55.3/3.0} & \textbf{47.5/3.1} & \textbf{71.1/3.9} \\
\hline
\end{tabular}
\caption{
Ablation studies on three variants of multi-scale processing in LinVT, evaluated under video benchmarks. Mipha-1.6B is used here. Among these variants, \emph{Multi-C} is chosen as our final design in LinVT.
}
\label{tab:suppl_ablation_multi}
\end{table*}
\begin{table*}[h!]
\centering
\resizebox{\textwidth}{!}{%
\begin{tabular}{c|ccccccccccc}
\hline
Text Conditions & {MVB} & {MMB-V} & {V-MME} & {LongVB} & {N-QA} & {EgoS} & {MVD-QA} & {MTT-QA} & {Act-QA} & {TGIF-QA}\\
\hline
& 61.8 & 1.19 & 47.6 / 48.9 & 49.1 & 70.4 & 54.9 & 70.1 / 3.6 & 54.6 / 2.9 & 47.0 / 3.1 & 70.4 / 3.7\\
\checkmark & \textbf{62.5} & \textbf{1.22} & \textbf{48.4 / 49.2} & \textbf{49.7} & \textbf{71.1} & \textbf{55.9} & \textbf{71.2 / 3.8} & \textbf{55.3 / 3.0} & \textbf{47.5 / 3.1} & \textbf{71.1 / 3.9} \\
\hline
\end{tabular}
}
\caption{Ablation studies on the effectiveness of text conditions in TTA. The image-LLM used in this experiment is Mipha-1.6B. Interaction with the text conditions boosts performance on several video benchmarks.}
\label{tab:suppl_textcondition}
\end{table*}

\section{Qualitative Analyses}
\label{sec:suppl_additional}
We present visualizations of selected tokens using the spatio-temporal significance scoring and selection mechanism, alongside examples of visual question answering (VQA) on long videos utilizing LinVT-InternVL2 (8B).

\begin{figure*}
    \centering
    \includegraphics[width=1.0\linewidth]{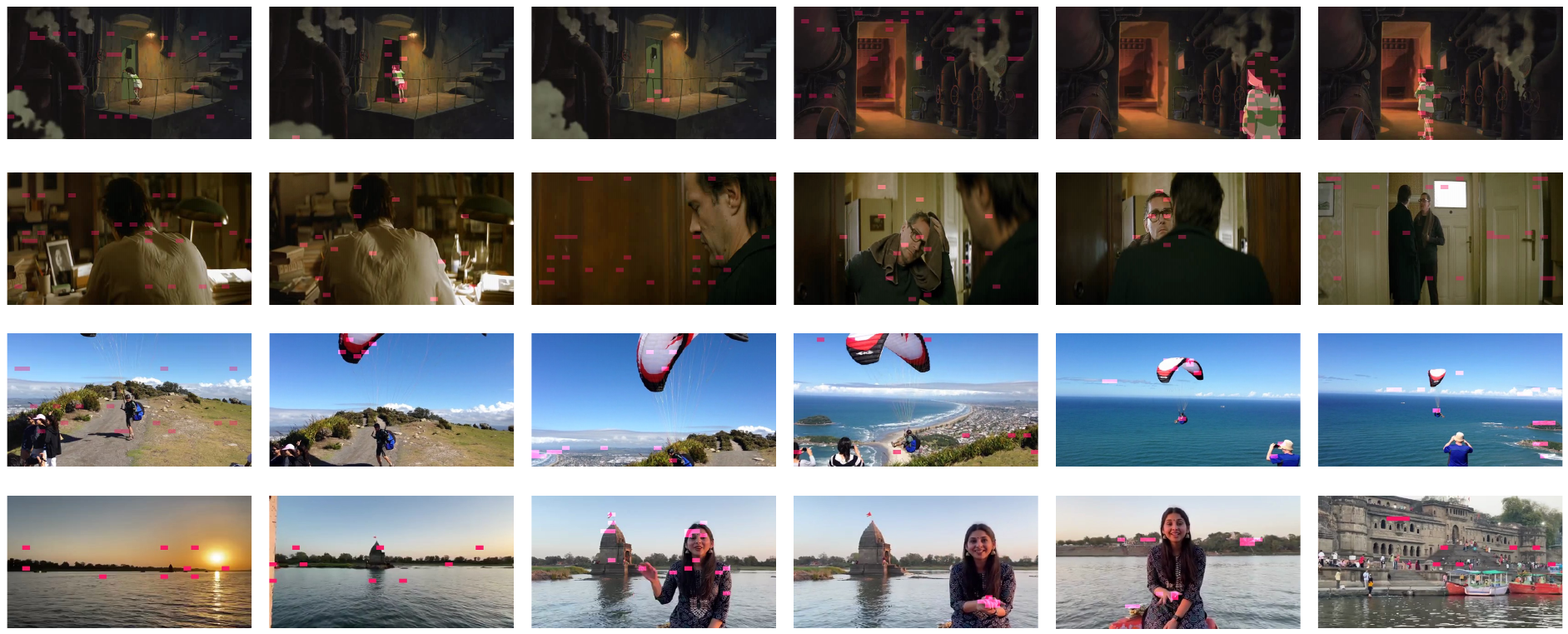}
    \vspace{0mm}
    \caption{Visualization of the patches corresponding to the selected tokens in video frames. The selection is achieved by the spatio-temporal significance scoring and top-{k} selection. The red patches in the image denote the selected tokens. This token scoring and selection mechanism directs the model's attention towards the most prominent objects, actions, or scenes within the video.
    LinVT-InternVL2-8B is used for this visualization.}
\label{fig:selector-suppl}
\end{figure*}

\paragraph{Image patches of the selected tokens.} 
In \cref{fig:selector-suppl}, we highlight the image patches corresponding to the tokens selected by the spatio-temporal significance scoring and selection mechanism. These visualizations reveal that the proposed mechanism effectively directs the model's attention toward the most prominent objects, actions, or scenes within the video.

\begin{figure*}
    \centering
    \includegraphics[width=1.0\linewidth]{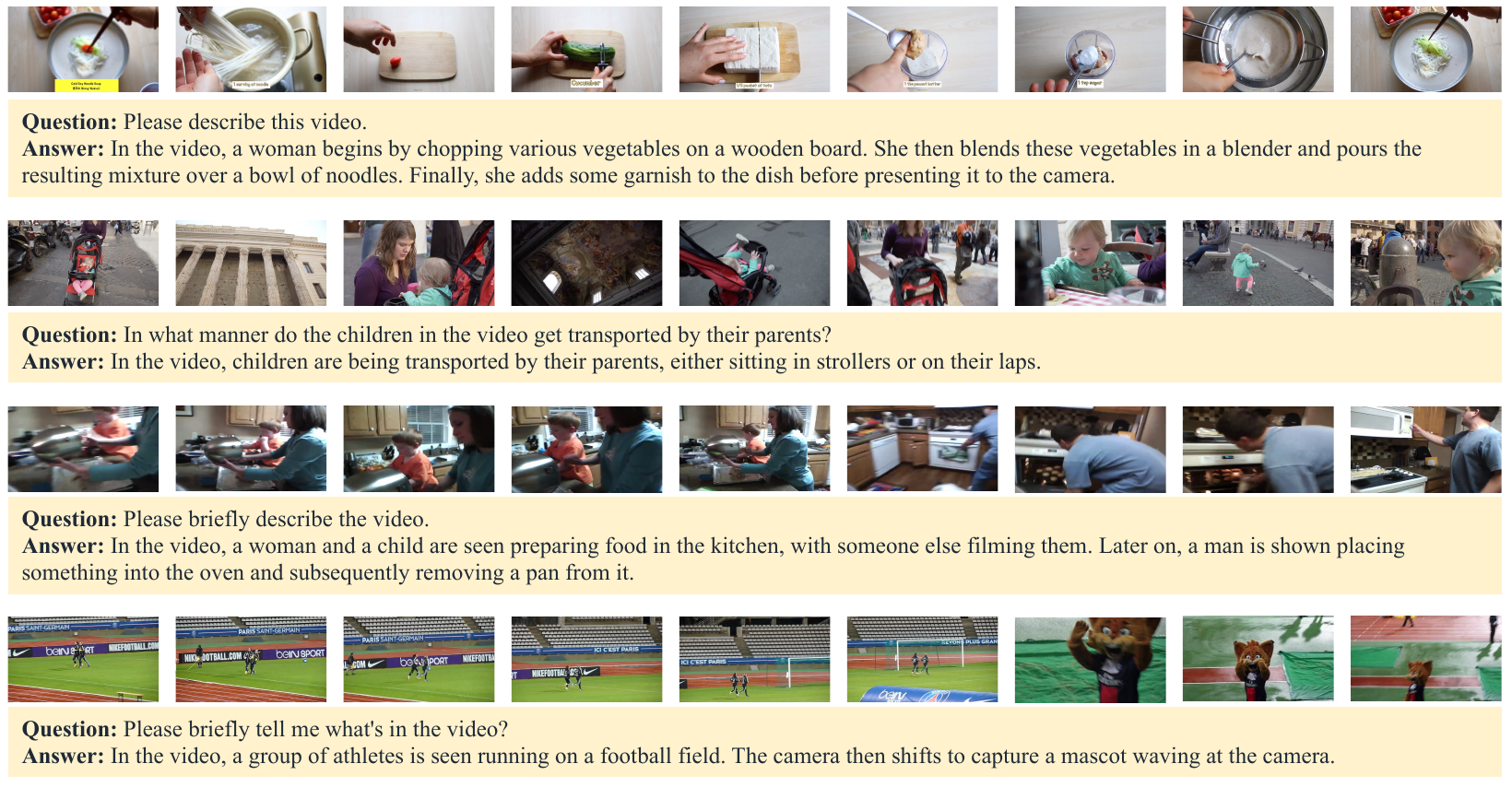}
    \vspace{0mm}
    \caption{Visualization of VQA on long videos. The captions generated by LinVT exhibit a greater focus on fundamental facts. The model's responses indicate that LinVT enhances the ability of video-LLM to understand videos in a more comprehensive manner, avoiding bias towards extraneous details such as background text.}
\label{fig:demo_general}
\end{figure*}

\paragraph{VQA on long videos.} As demonstrated in \cref{fig:demo_general}, the incorporation of the proposed LinVT modules enables Video-LLM to generate answers that are more factual and less biased by irrelevant details.

\end{document}